# Kinematics analysis and three-dimensional simulation of the rehabilitation lower extremity exoskeleton robot

Jiangcheng Chen, Xiaodong Zhang and Lei Zhu
School of Mechanical Engineering
Xi'an Jiaotong University
Xi'an, P.R. China
e-mail: amct@mail.xjtu.edu.cn

*Abstract*— The kinematics recursive equation was built by using the modified D-H method after the structure of rehabilitation lower extremity exoskeleton was analyzed. The numerical algorithm of inverse kinematics was given too. Then the three-dimensional simulation model of the exoskeleton robot was built using MATLAB software, based on the model, 3D Reappearance of a complete gait was achieved. Finally, the reliability of numerical algorithm of inverse kinematics was verified by the simulation result. All jobs above lay a foundation for developing a three-dimensional simulation platform of exoskeleton robot.

## I. INTRODUCTION

Rehabilitation robot which introduced robot technology into rehabilitation engineering, reflect the perfect combination of rehabilitation medicine and robotics. It is currently a hot research field of robotics. Under the control of gait simulation control system, lower limb rehabilitation robot can simulate the motion of normal human walking gesture which help the patients who has a walking disability doing rehabilitation training, help them to recover their nerve and muscle function. Compared to traditional artificial lower limbs rehabilitation training method, the electromechanical device assists patients to complete repetitive walking motion, can freed the trainers from onerous task, shows a strong advantage. Presently, lower limb rehabilitation robot can be divided into two kinds: leg powered exoskeleton and plantar driven pedal type [1-3].

The capability of exoskeleton's gait simulation affect the patient's rehabilitation effect directly, it depends on the reasonable structure design and a gait planning and control method. As for the design of structure, the exoskeleton should be anthropomorphic and ergonomic, not only in shape but also in function, so that the exoskeleton can produce the same pose as human lower limb can. At the aspect of gait generation, effective gait planning and control algorithms are needed. However, exoskeleton system is a high-level, strongly coupled non-linear, multi-degree of freedom, complex mechanical system, which brings difficulties to the system design and motion control. Modeling and simulation provide effective adjunct to the development of the system, the simulation results is able to verify the correctness of system design, the validity of the motion planning and control algorithms. In this paper, we conduct the exoskeleton robot's kinematics analysis, and achieve the 3D visualization of movement under MATLAB environment .It lays a foundation for the development of the entire exoskeleton system simulation platform.

## II. RELATED WORKS

Simulation is an important part in the robotic research. The most successful example is that the Japanese scholars, Hirohisa Hirukawa, Fumio Kanehiro, etc. developed two biped robots, HRP-1 and HRP-2, by using of 3D simulation platform [4].A strict mathematical model and physical model should be established before building the virtual prototype model of the mechanical system. The lower exoskeleton is a complex and multi-degree of freedom system. In order to facilitate the analysis, it is necessary to simplify the dynamical structure of the robot. The early simplified models are planner link model, such as five-linkage model, seven-linkage model and nine-linkage model [5-6]. The coupling between the forward movement and lateral movement is neglect in these models. Another successful model relatively is the 3D inverted pendulum [7]. But this kind of model is not suitable for gait planning. Further more, in the simulation of the exoskeleton robot, several modeling and simulation Software were used which lead to that large number of data exchange was exist during the simulation. In this paper, a 3D linkage model is used and all simulation is carried out in MATLAB.

## III. EXOSKELETON STRUCTURE ANALYSIS

The design of exoskeleton robot follows the principle of structure and function bionic, it comprised of two booster mechanical legs, two feet which pass gravity to the ground, and waist ring whose width was adjustable, in which the waist ring is connect to the balance support. Mechanical legs are made up with hip module, knee module, the thigh link module and calf link module. The typical representative of exoskeleton rehabilitation robot's are the Swiss walking rehabilitation robot system LOKOMAT and the Dutch lower extremity power bones LOPES. Each leg of LOKOMAT exoskeleton system includes two revolute joints, located in hip and knee. The movement of joints was achieved by ball screw with independent motor. Ankle is controlled by the treadmill belt in support phase of gait. By contrast, LOPES provides more degrees of freedom, there are two degrees of freedom in hip: flexion / extension, outreach / adduction, one degree of freedom in knee: flexion / extension, one degree of freedom in pelvis: up / down. There are 7 degrees of freedom in total [8-9].

However, by the mechanism analysis of human lower limb, we can find that the human lower limbs contain 12 degrees of freedom, which are the hip's flexion / extension, outreach / adduction and internal rotation / external rotation, knee's flexion / extension movement, ankle's flexion / extension and outreach / adduction. Therefore, in order to simulate a more suitable three-dimensional gait for human motion, we must increase more degrees of freedom in the existing rehabilitation exoskeleton structure.

## IV. KINEMATICS MODELING AND ANALYSIS

The kinematics analysis of exoskeleton robot is the basis of the movement position and posture analysis and gait planning, which includes the forward kinematics and the inverse kinematics. Forward kinematics is used to find the position, velocity, acceleration and attitude of each link in the base coordinate system in the case of giving the relative position and joint variable parameters. Forward kinematics will be used in the calculation of the center of gravity and the detection of contact state between robot and the environment. What's more, forward kinematics is also the basics of realizing the visualization of three-dimensional simulation. D-H method and the modified D-H method [10-11] are the commonly used methods in forward kinematics calculation. On the contrary, the inverse kinematics is needed if the known conditions are the pose of hindquarters and feet, we can calculate the joint variable parameters with inverse kinematics analysis. The calculation methods of inverse kinematics can be divided into analytical solution, the geometric solution and the numerical solution. Pieper criteria should be met in the geometric structure while using analytical method, that is, only three adjacent joint axes intersect at one point or three adjacent joint axes parallel to each other, can geometric closed solution be used for solving the inverse kinematics, What's more, the solving progress would become more complex with the degrees of freedom increasing. By contract, there has no such a limitation to the numerical solution.

### A. Establishment of Coordinate System

The exoskeleton robot includes 12 degrees of freedom. Due to so many degrees of freedom and the coupling exists in movements, it is difficult to describe its position and orientation with a base coordinate system, local dynamic coordinate system for each degree of freedom is necessary. For the lower limbs rehabilitation robot, we can consider the body as the base because it keeps stationary in movements. So we can establish the coordinate system by using the modified D-H method, which is shown in figure 1. The local coordinate system established by using the modified DH method will not change while the rod is changing, that is the different between the modified D-H method and conventional D-H method. This different can eliminate the accumulated error in the calculation process.

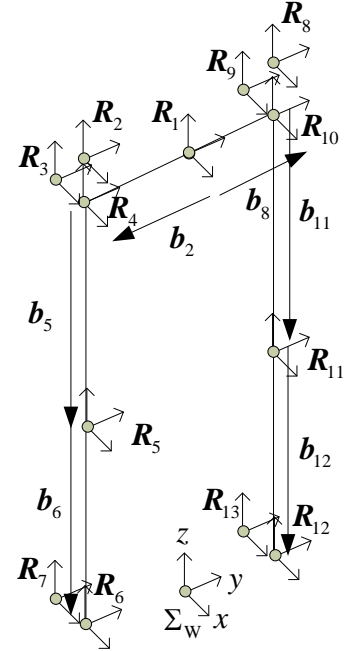

Figure 1 Establishment and the relative position relationship of the coordinate system

In Figure 1, world coordinate system $\Sigma_W$ is in the middle of two foot, $R_i$ ($i=1,2\text{L }13$) represent local coordinate system of each joint, $b_i$ ($i=1,2\text{L }13$) represent the position vector between the two coordinate systems. The axis of all the local coordinate system parallel correspondingly to the axis of world coordinate system in the initial state. Through the establishment of the coordinate system, the exoskeleton system can be seen as two mechanical chains whose base is the body.

### B. Forward kinematics

The calculation process of forward kinematics can be divided into two steps: 1) calculate the homogeneous transformation matrix between the connected coordinate system; 2）Kinematical calculation by the chain multiplication of transform matrix. According to the coordinate system established above, the homogeneous transformation matrix between coordinate system between the $\Sigma_i$ and $\Sigma_j$ can be expressed as [7]

$$^iT_j = \begin{bmatrix} e^{\hat{a}_j q_j} & b_j \\ 0_{1\times 3} & 1 \end{bmatrix} \quad (1)$$

Where, $e^{\hat{a}_j q_j} = I + \hat{a}_j \sin q_j + \hat{a}_j^2(1-\cos q_j)$ is the attitude Matrix, and $I$ is a unit vector, $q_j$ is joint rotation angle, $\mathbf{a}_j = [a_x, a_y, a_z]$ is rotary axis vector, $\hat{\mathbf{a}}_j$ is the antisymmetric matrix of $\mathbf{a}_j$, can be expressed as follows

$$\hat{\mathbf{a}}_j = \begin{bmatrix} 0 & -a_z & a_y \\ a_z & 0 & -a_x \\ -a_y & a_x & 0 \end{bmatrix} \quad (2)$$

Assuming the homogeneous transformation matrix of coordinate system $\Sigma_i$ relative to the world coordinate system is

$$^w\mathbf{T}_i = \begin{bmatrix} \mathbf{R}_i & \mathbf{p}_i \\ 0_{1\times 3} & 1 \end{bmatrix} \quad (3)$$

Then the homogeneous transformation matrix of coordinate system $\Sigma_j$ relative to the world coordinate system can be got by using chain multiplication

$$^w\mathbf{T}_j = {}^w\mathbf{T}_i\,{}^i\mathbf{T}_j \quad (4)$$

Finally, the absolute position and posture can be obtained by (1) (3) and (4)

$$\mathbf{p}_j = \mathbf{p}_i + \mathbf{R}_i \mathbf{b}_j \quad (5)$$

$$\mathbf{R}_j = \mathbf{R}_i e^{\hat{\mathbf{a}}_j q_j} \quad (6)$$

If the body's position and orientation are known and all the joint angles are given, we can calculate the position and attitude of each link according the formulas of (5) and (6).

### C. Inverse kinematics

In the exoskeleton robot, inverse kinematics is a method for solving each angle under the condition that the position and orientation of the torso and feet are known. As the exoskeleton robot has 12 degrees of freedom, it is difficult to obtain the analytical solution. Numerical solution method is flexible and has no limit. In this paper, we use numerical method which is based on forward kinematics calculation. The principle for the inverse kinematics is: Giving an initial value of articulation angle, calculate the foot posture currently through the forward kinematics, compared the results with the target value, and change the angle value with the angle correction, and then recycled, until the error is within a certain range. As shown in Figure 2.

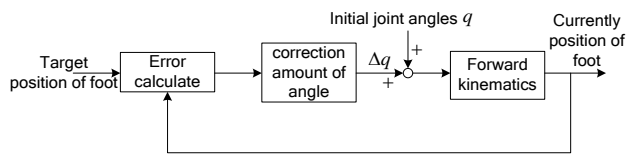

Figure 2 the principle of inverse kinematics using numerical method

## V. THE THREE-DIMENSIONAL KINEMATICS SIMULATION

### A. Three-dimensional Simulation Model

Due to the complexity of exoskeleton system, the kinematics equations have the characteristics of nonlinear and strong coupling, the kinematics calculation results were shown with the form of curves. The combination of a variety of simulation methods can achieve a three-dimensional visualization [12-13], but it needs a large amount of data exchange, the accumulation of data error between different software will affect the simulation results. Using MATLAB software, and using the data structure of the family spectrum, we built a 12-DOF exoskeleton system model, shown in Figure 3.

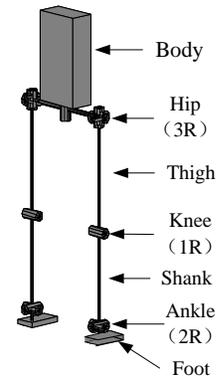

Figure3 Three-dimensional Model of exoskeleton

### B. Gait visual reproduction

In order to verify the validity of the forward kinematics, a kinematics simulation was done with the clinical gait data which is given by the Polytechnic University of Hong Kong [14], as shown in Figure 4. Meanwhile, 10 poses of the robot in the process of gait movement were taken out, shown in Figure 5, all the axis unit is cm.

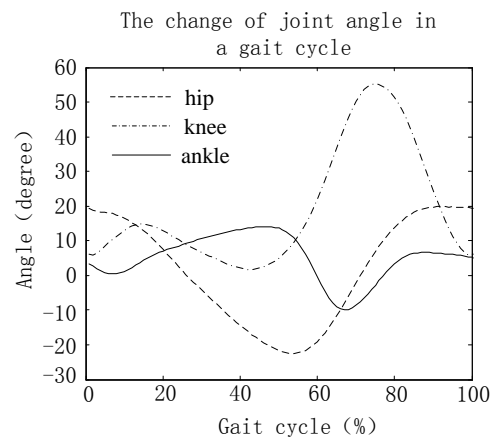

Figure 4.The Hong Kong Polytechnic University clinical gait data[9]

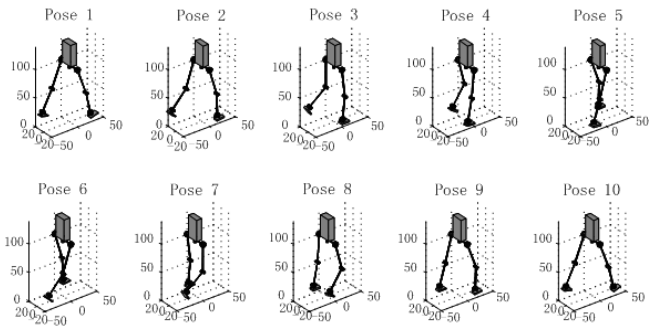

Figure 5 10 poses of the robot in a gait cycle

*C. Inverse kinematics algorithm simulation*

According to the numerical method of inverse kinematics in 3.3 and with the foot trajectory in a gait cycle calculated through the forward kinematics method proposed in 3.2, the change trajectory of each joint were got by simulation in the MATLAB, shows in figure 6(dotted lines). In order to contrast, the measured joint trajectory curves are also given in the figure. According to the comparison, we can find out that the numerical method of inverse kinematics can obtain satisfactory results.

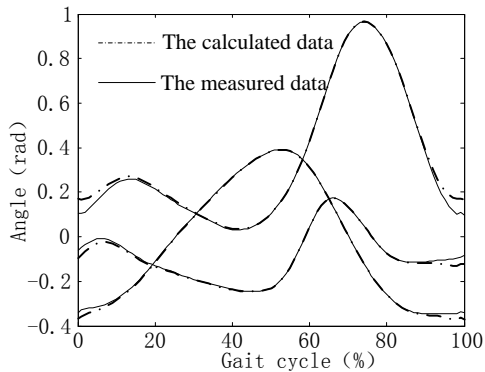

Figure 6 Numerical results of the inverse kinematics

## VI. CONCLUSION

Rehabilitation exoskeleton robot, as a new type of medical equipment and typical man-machine integration equipment, is a hot research field in robotics. This paper discussed the structure of rehabilitation exoskeleton robot, deduced the forward kinematics calculation equations and inverse kinematics numerical methods. The simulation model of the robot proposed in this paper can be used in the development of a Rehabilitation exoskeleton robot, through this model a three-dimensional visualization of movement can be realized, if the dynamics progress is added into the model too, we can develop a complete 3D simulation platform.

Therefore, in order that the kinematics and dynamic progress are both involved in the simulation, the dynamics would be analyzed and added into the model in our future work. And our ultimate target is that the simulation model can be driven not only by kinematics parameters but also by dynamics parameters. As a result, this model can be used for the validation of walking planning and control algorithm.

## VII. ACKNOWLEDGMENT

The authors are very grateful for the support provided by the National Natural Science Foundation of China (GrantNo.51275388).